\documentclass[10pt,twocolumn]{article}

\usepackage[utf8]{inputenc}
\usepackage[T1]{fontenc}
\usepackage[margin=0.85in,columnsep=0.28in]{geometry}
\usepackage{amsmath,amssymb}
\usepackage{graphicx}
\usepackage{booktabs}
\usepackage{multirow}
\usepackage{enumitem}
\usepackage{xcolor}
\usepackage{caption}
\usepackage{microtype}
\usepackage{tikz}
\usepackage{pgfplots}
\pgfplotsset{compat=1.17}
\usetikzlibrary{patterns}
\usepackage{placeins}
\usepackage{balance}
\usepackage[colorlinks=true,linkcolor=black,citecolor=black,urlcolor=blue]{hyperref}

\makeatletter
\renewcommand\section{\@startsection{section}{1}{\z@}%
  {-3.0ex \@plus -1ex \@minus -.2ex}{1.5ex \@plus .2ex}%
  {\normalfont\large\bfseries\raggedright}}
\renewcommand\subsection{\@startsection{subsection}{2}{\z@}%
  {-2.6ex \@plus -1ex \@minus -.2ex}{1.1ex \@plus .2ex}%
  {\normalfont\normalsize\bfseries\raggedright}}
\makeatother

\definecolor{dalblue}{RGB}{21,64,164}   
\definecolor{dalpale}{RGB}{176,198,235} 
\definecolor{dalink}{RGB}{0,0,0}        

\title{\bfseries\Large Does a Modern-Handwriting Warm-Up Help\\ Historical Arabic OCR?\\[0.35em]
\normalsize A Reproducible, Compute-Matched Evaluation on Muharaf and KHATT}

\author{%
\normalsize Sumaih Almarshad, Maram Alamri, Ph.D., Dona Aloraini, Fares Altuwaim,\\
\normalsize AlJawharh AlOtaibi, Reem Alyabis, Rayah Aldawsari\\[0.35em]
\normalsize Dal Research Team, Riyadh, Saudi Arabia\\
\normalsize \texttt{\{sumaih, maram, dona, fares, aljawharh, reem, rayah\}@daldata.ai}}

\date{August 18, 2026}

\begin{document}
\maketitle

\begin{abstract}
\noindent
Whether an intermediate stage of modern Arabic handwriting helps or hurts historical Arabic
handwritten text recognition (HTR) is usually decided from one implementation and one
comparison, which is too thin a basis for a claim in either direction. We test how stable that
effect is by running the same nominal ablation four times inside a single project, letting the
base checkpoint, encoder-freezing strategy, epoch budget, numerical precision, and
learning-rate schedule vary as they naturally did during development, while holding the
normalization policy, the scorer, and the interval estimation fixed. Each run compares
intermediate training on modern handwriting (KHATT) followed by fine-tuning on historical
manuscripts (Muharaf) against fine-tuning on Muharaf directly. Across the four runs the
estimated effect swings from $-17.64$ to $+14.52$ CER points and reverses sign. That range is
not evidence of a chaotic ablation: the two extremes are exactly the two runs with an
identifiable confound. One lowers the target-stage learning rate fivefold in the staged arm;
the other builds on a checkpoint of undisclosed provenance. The two runs without either
problem land at $-0.25$ and $+0.94$, that is, no effect. The lesson is that a tight interval
from one implementation says nothing about the next. We then run a compute-matched experiment
with identical warm-up and target-stage budgets over three seeds. There, KHATT warm-up is
$+2.42$ CER points worse than a compute-matched same-domain control (paired seed-level 95\%
interval $[+0.60, +4.25]$); measured against a disjoint slice of real target data, the part of
that gap specific to the modern-handwriting domain is only about 0.6 points. We read this as a
small negative effect under this configuration, not a universal result. We release a
SaudiHeritage-OCR package containing the shared normalizer and interval scorer, a verified
KHATT coded-label decoder, experimental manifests, VLM baselines, and an edition-alignment
protocol for future Saudi heritage data, so the result can be checked independently. The single
Al-Mahd inscription line is held strictly out and is not offered as a benchmark. The
contribution is a controlled transfer-learning result together with the infrastructure to test
how far it travels.
\end{abstract}

\noindent\textit{All authors contributed equally to this work.}

\vspace{0.6em}
\noindent\footnotesize\textbf{Index Terms:} Arabic handwritten text recognition, OCR, historical
manuscripts, epigraphy, vision-language models, LoRA, data schedule, intermediate-task transfer,
reproducibility, confidence intervals, Saudi heritage, implementation variance, ablation study.
\normalsize

\section{Introduction}

In June 2026 the Saudi Heritage Commission announced 1,774 archaeological finds from the
Al-Mahd survey in the Madinah region, among them 461 Islamic and 34 Thamudic inscriptions and a
rare early Hijazi rock inscription invoking the name of the Caliph `Umar ibn al-Khattab
[1, 2, 3]. Discoveries like these are arriving faster than they can be read. Backed by Saudi
Vision 2030 work on cultural-heritage preservation and digitization, the documented corpus of
historical Arabic text keeps growing, and its scale and variety strain any transcription effort
that runs through epigraphists and codicologists by hand.

Automated HTR is one way to relieve that pressure. Transformer-based systems now reach
single-digit character error rates (CERs) on historical Arabic manuscripts [4], and large
vision-language models (VLMs) transcribe Arabic well without task-specific training [5, 6]. For
anyone adapting such a model to Saudi heritage material, a concrete question about resources
follows: given limited annotation and compute, does an intermediate stage on a larger, more
available corpus of modern Arabic handwriting help once the model is fine-tuned on a smaller
historical target?

The transfer-learning literature gives no blanket answer. An intermediate stage helps in some
settings and hurts in others, depending on how the intermediate and target domains relate and on
the training configuration around them [19, 20, 21, 22]. Yet conclusions on this question in
Arabic HTR usually rest on a single reported ablation, which leaves open whether the observed
effect belongs to domain transfer or to one implementer's choices.

\begin{figure}[t]
\centering
\begin{tikzpicture}
\begin{axis}[
  width=\linewidth, height=4.8cm,
  xmin=20, xmax=95,
  xtick={20,40,60,80},
  ytick={1,2,3,4},
  yticklabels={I1,I4,I3,I2},
  ymin=0.4, ymax=4.6,
  xlabel={Normalized CER (\%) on Muharaf, lower is better},
  tick label style={font=\footnotesize},
  label style={font=\footnotesize},
  axis lines*=left,
  y axis line style={draw=none},
  ytick style={draw=none},
]
\addplot[dalblue,thick] coordinates {(45.86,4) (28.22,4)};
\addplot[only marks,mark=o,mark size=2.6pt,dalblue,line width=0.9pt] coordinates {(45.86,4)};
\addplot[only marks,mark=*,mark size=2.6pt,dalblue] coordinates {(28.22,4)};
\node[anchor=west,font=\scriptsize] at (axis cs:48.5,4) {$-17.64$};
\addplot[dalink,thick] coordinates {(82.88,3) (82.63,3)};
\addplot[only marks,mark=o,mark size=2.6pt,dalink,line width=0.9pt] coordinates {(82.88,3)};
\addplot[only marks,mark=*,mark size=2.6pt,dalink] coordinates {(82.63,3)};
\node[anchor=west,font=\scriptsize] at (axis cs:85.0,3) {$-0.25$};
\addplot[dalink,thick] coordinates {(77.51,2) (78.45,2)};
\addplot[only marks,mark=o,mark size=2.6pt,dalink,line width=0.9pt] coordinates {(77.51,2)};
\addplot[only marks,mark=*,mark size=2.6pt,dalink] coordinates {(78.45,2)};
\node[anchor=west,font=\scriptsize] at (axis cs:80.7,2) {$+0.94$};
\addplot[dalblue,thick] coordinates {(42.81,1) (57.33,1)};
\addplot[only marks,mark=o,mark size=2.6pt,dalblue,line width=0.9pt] coordinates {(42.81,1)};
\addplot[only marks,mark=*,mark size=2.6pt,dalblue] coordinates {(57.33,1)};
\node[anchor=west,font=\scriptsize] at (axis cs:59.5,1) {$+14.52$};
\end{axis}
\end{tikzpicture}
\caption{The central result. Four separately configured implementations: one project, one month,
the same two corpora, one normalizer, one scorer, one question. Does a KHATT warm-up before Muharaf
fine-tuning help? Each row runs from the single-stage result (hollow) to the staged result
(filled); $\Delta$CER is at right. Rows are sorted by $\Delta$CER, matching
Table~\ref{tab:central}. The effects span 32.16 points and the sign flips. Read carefully, the
spread is driven by the two rows with identifiable confounds. I1 lowers its target-stage learning
rate fivefold in the staged arm, and I2 builds on a checkpoint of undisclosed provenance. I3 and I4
(black) agree on ``no effect.'' Viewed in isolation, individual rows would support materially
different conclusions.}
\label{fig:central}
\end{figure}
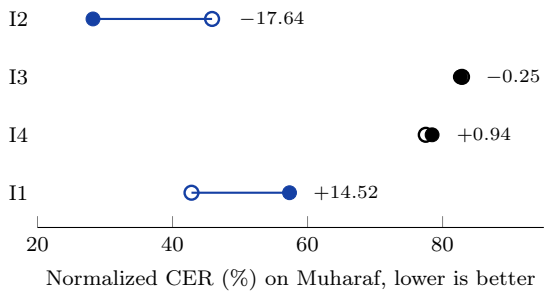

We look at that question through four implementations of the same nominal ablation. They were
not built as a replication study; they came out of parallel development and differed in base
checkpoint, encoder-freezing strategy, epoch budget, numerical precision, and learning-rate
schedule, which are the ordinary degrees of freedom anyone adapting an HTR system meets in
practice. We use \emph{independent} in exactly that sense: separately configured runs, sharing
the datasets and the final evaluation code but not a common checkpoint, schedule, freezing
policy, precision, or manifest. What stayed fixed was the comparison itself, KHATT warm-up then
Muharaf fine-tuning against direct fine-tuning on Muharaf, and the evaluation, which passed
every run through one normalizer and one interval-based scorer. Even so, the measured effect
moved across roughly 32 CER points and changed sign (Fig.~\ref{fig:central}).

It is tempting to read that spread as an inherently unstable ablation. The more accurate reading
is narrower, and we prefer it because it is what the numbers support. The two rows at the
extremes are the two runs with a concrete problem. One (I1) trained the staged arm at a learning
rate five times lower than its own baseline, a stabilization choice that suppresses adaptation
whatever came before. The other (I2) builds on a community checkpoint of undisclosed provenance,
so its absolute level cannot be trusted. The two runs with neither problem, I3 and I4, both with
a frozen encoder, sit at $-0.25$ and $+0.94$: essentially no effect. So the matrix does not show
a coin-flip ablation. It shows that different configurations give different answers, that two of
those differences are explained, and, most usefully, that a precise interval inside any one run
is silent about the others.

That last point is the methodological core, and it is worth stating plainly because the field's
current practice hides it. In I1, KHATT warm-up produced a $+14.52$-point mean increase in CER
(paired-line bootstrap point estimate $+14.58$), with an interval of $[+13.97, +15.19]$ that
excludes zero. Nothing about that interval is wrong. It quantifies uncertainty over the lines
sampled within that run, and it does so correctly. What it cannot see is the variation introduced
by a different but equally reasonable training configuration. A clean estimate from one
implementation can therefore be an incomplete, even misleading, basis for a broad claim whenever
implementation-level variation is large, as it is here.

We treat that cross-implementation variability as a finding in its own right, then ask whether
the direction of the effect can be pinned down under control. A fifth run was built for that:
warm-up and target-stage budgets fixed, three shared seeds, domain of the warm-up corpus the only
quantity allowed to move. There, KHATT warm-up was worse than a compute-matched same-domain
control by 2.42 CER points, consistently across seeds, which is evidence for a small negative
effect within this configuration. It is not a universal effect, and, as Section~\ref{sec:matched}
shows, most of even that small gap disappears once the comparison is made against a disjoint
slice of real target data rather than against extra passes over the target.

The broad implication is about evidence, not about Arabic OCR. A transfer-learning conclusion in
this setting should not rest on the confidence interval of a single implementation; evaluation
uncertainty and implementation variability are different things and deserve to be reported apart.
Keeping them apart gives a more honest reading of whether modern Arabic handwriting is a useful
intermediate domain for historical HTR, and it leaves behind a reproducible way to test such
strategies on Saudi heritage material.

\subsection*{Contributions}

\begin{itemize}[leftmargin=1.2em,itemsep=0.35em,topsep=0.35em,parsep=0pt]
\item \textbf{Evidence on implementation sensitivity.} Four separately configured runs of the
same nominal KHATT$\rightarrow$Muharaf ablation give effects from $-17.64$ to $+14.52$ CER
points, including a sign reversal. The spread is driven by the two runs with identifiable
confounds; the two clean runs agree on no effect. The general lesson holds regardless: one
implementation's interval does not bound what a re-implementation will find.

\item \textbf{A compute-matched follow-up.} A fifth experiment fixes the warm-up and target-stage
budgets, corpus size, optimizer, target learning rate, and shared seeds. Across three seeds KHATT
warm-up is $+2.42$ CER points worse than the compute-matched same-domain control
($[+0.60, +4.25]$); against a disjoint real-target slice the domain-specific part is about 0.6
points. We frame it as a small negative effect for this reader and configuration, not a general
claim.

\item \textbf{Reproducible infrastructure.} We release the shared normalizer and scorer, the
interval and paired-difference code, a verified KHATT coded-label decoder, the manifests, and the
VLM and edition-alignment utilities. The Al-Mahd material stays held out and is not presented as
a powered benchmark.
\end{itemize}

\section{Related Work}

\subsection{From traditional to neural OCR}

Classical OCR leaned on preprocessing, character segmentation, and template matching [8], and
early Arabic systems ran a sequential pipeline of segmentation, feature extraction, and
recognition against handcrafted models [8]. Arabic makes segmentation brittle: cursive joins,
contextual letter shapes, ligatures, and diacritics resist clean cuts, and the problem worsens on
handwritten and historical pages [8]. Deep learning shifted the field onto learned
representations, first CNN feature extractors with recurrent sequence models under Connectionist
Temporal Classification and later attention, which raised recognition of printed and handwritten
Arabic substantially [8]. Al-homed et al.~[9] paired Faster R-CNN detection with LSTM recognition
on historical Arabic catalog cards, though CNN-LSTM models stay fragile on badly degraded
material. Comparisons of classical pipelines with generative, AI-assisted OCR repeat the theme:
the winner depends heavily on the material and the configuration [7].

\subsection{Transformer and vision-language OCR}

TrOCR [10] set the template for end-to-end Transformer OCR, joining a pretrained Vision
Transformer encoder [11] to a pretrained Transformer decoder through cross-attention [12].
HATFormer [4] carried the architecture onto historical Arabic manuscripts and reported 8.6\% CER
on Muharaf [14], the strongest published number on that benchmark and our reference point. VLMs
push the idea further with large multimodal pretraining; KITAB-Bench [6] tracks their growing but
still limited command of specialized historical Arabic. Low-rank adaptation [15] keeps
fine-tuning of such models affordable and underlies our VLM reader.

\subsection{Intermediate-stage transfer, and why it resists measurement}

An intermediate stage before target fine-tuning can help or hurt, and which one depends on how
intermediate and target relate. Supplementary training on intermediate tasks lifts some targets
and lowers others, with transfer governed by task and domain similarity rather than by extra data
as such [19, 20, 21]. Domain- and task-adaptive continued pretraining tells the same story from
the other side: in-domain continuation reliably helps, off-domain continuation helps little or
hurts [22]. Sequential fine-tuning also risks catastrophic forgetting and negative transfer [23].
The same literature explains why the effect is hard to isolate: inserting a stage changes compute,
initialization, and domain at once. Our results add a fourth factor that is rarely controlled,
namely the optimization settings of the target stage, which implementers quietly adjust to make a
staged run train cleanly. Section~\ref{sec:fourimpl} shows that this adjustment can exceed the
effect under study.

\subsection{Reproducibility of ablations}

Reported gains in machine learning often fail to survive a change of implementation, tuning
budget, or seed, and comparisons without matched tuning tend to favor whichever condition got
more attention [24, 25, 26]. Our result is a domain-specific case: a schedule ablation in Arabic
HTR whose sign is not stable across implementations of one nominal protocol.

\subsection{Arabic HTR resources and AI for heritage text}

KHATT [13] is the standard benchmark for modern Arabic handwriting, drawn from roughly a thousand
writers. Muharaf [14] offers 24,495 expertly transcribed historical text-line images, the closest
public dataset to historical Saudi archival material. KITAB-Bench [6] evaluates VLMs on Arabic OCR
and finds open-source models trailing proprietary ones zero-shot. None hold Saudi epigraphic
material, whose writing surfaces, weathering, and broken characters motivate our held-out
inscription case. Ithaca [16] showed that deep networks can restore missing characters, attribute
provenance, and date damaged ancient Greek inscriptions, and generative heritage-restoration work
has broadened since [17]. In Saudi Arabia, efforts led by SDAIA [18] and the Libraries Commission
have grown digital preservation but have emphasized digitization and cataloguing over automated
reading. We aim at that gap, contributing the evaluation infrastructure needed to judge automated
readers on Saudi heritage material.

\section{Data}
\label{sec:data}

Table~\ref{tab:data} lists the three corpora and the role each plays. We keep those roles strictly
apart: Muharaf is the historical target, KHATT is used only for intermediate training, and Al-Mahd
is reserved for out-of-domain qualitative inspection.

\begin{table*}[t]
\centering
\caption{Data sources and roles. Only public corpora are used for training; Al-Mahd is never
trained on. Counts are train / val / test where applicable.}
\label{tab:data}
\small
\begin{tabular*}{\linewidth}{@{\extracolsep{\fill}}lllr@{}}
\toprule
Source & Surface & Role & Lines \\
\midrule
KHATT [13]      & modern ink      & warm-up          & 9,497 / 1,901 \\
Muharaf [14]    & historic ink    & train + bench.   & 22,091 / 1,069 / 1,334 \\
Al-Mahd (ours)  & inscribed stone & test-only, qual. & 1 \\
\bottomrule
\end{tabular*}
\end{table*}

\textbf{Muharaf} [14] holds 24,495 line images from historical Arabic manuscripts, each with an
expert-verified transcription. To stay comparable with prior work we keep the original Hugging
Face train/validation/test partition and drop samples with empty transcriptions, leaving 22,091
training, 1,069 validation, and 1,334 test lines. The frequently cited 22,092 is the pre-exclusion
figure; one training instance has an empty transcription. The public release carries no writer
identifiers, so we cannot verify that the partition is writer-independent; rather than invent a
new split we keep the original and note the limitation in Section~\ref{sec:limits}. One run (I4)
built its own manifest and resplit the corpus into 20,819 training, 2,449 validation, and 1,226
test lines; we treat that resplit as one of the implementation differences under study, and note
in Section~\ref{sec:limits} that it makes I4's two conditions internally comparable but its
absolute level not comparable to the other runs.

\textbf{KHATT} [13] is a large benchmark for modern Arabic handwriting, used here only as the
intermediate corpus and never for final evaluation. The commonly mirrored release stores
transcriptions in a coded-column format rather than as Arabic text. To make it reproducibly usable
we wrote a decoder against the official KHATT v1.0 lookup table; it covers every code in the
training and validation CSVs (100\% on both) and yields 9,497 training and 1,901 validation lines.
Because Arabic reads right to left, we also check label ordering against the line images during
preprocessing. The decoder ships with the study so that experiments on this KHATT distribution can
be reproduced. The compute-matched follow-up (I5) uses KHATT only as the intermediate domain, with
the warm-up and target-stage step budgets, warm-up corpus size, and optimizer held constant across
arms; the full design is in Sections~\ref{sec:matrix} and~\ref{sec:matrixsetup}.

\textbf{Al-Mahd inscriptions.} This material is kept for qualitative, out-of-domain inspection
only. We hold out line crops from early Islamic Hijazi inscriptions documented in the 2026 Al-Mahd
survey, with readings taken from the corresponding published epigraphic edition. None of it is
used for training, fine-tuning, validation, or adaptation. The one line reported here has a
reference reading produced by expert post-editing of a model prediction. Because post-editing a
model's own output can bias evaluation, since the corrected reading may echo the model's errors,
and because the set holds a single line ($N = 1$), we treat the result as qualitative and make no
quantitative claim about manuscript-to-stone transfer from it.

\section{Methodology}

\subsection{Normalization policy}

\begin{table*}[t]
\centering
\caption{TrOCR-family implementations, all scored under the single normalization policy of
Sec.~\ref{sec:norm} and the same scorer. I1 to I4 are the four pilots. They are not a controlled
replication but the ordinary spread of choices four implementers made, which is precisely what
this study measures. I5 is the compute-matched follow-up reader (Sec.~\ref{sec:matrix},
\ref{sec:matrixsetup}); unlike I1 to I4 it is step-budgeted (Table~\ref{tab:hyper}) with a single
learning rate shared across all arms and stages, so its epoch and M1/M2 columns do not apply
($\dagger$).}
\label{tab:impl}
\small
\begin{tabular*}{\linewidth}{@{\extracolsep{\fill}}lllcccc@{}}
\toprule
 & Base checkpoint & Encoder & Epochs & Batch & Precision & Target-stage LR (M1 / M2) \\
\midrule
I1 & \texttt{RayR1/trocr-base-arabic-hw} & trainable & 5  & $8 \times 2$ & fp16 & $5\times10^{-5}$ / $1\times10^{-5}$ \\
I2 & \texttt{RayR1/trocr-base-arabic-hw} & trainable & 8  & 16 & fp16 & $5\times10^{-5}$ / $5\times10^{-5}$ \\
I3 & ViT-B/16-384 + AraBERT              & frozen    & 15 & 16 & fp32 & $5\times10^{-5}$ / $5\times10^{-5}$ \\
I4 & ViT-B/16-384 + AraBERT              & frozen    & 15 & 8  & fp32 & $3\times10^{-5}$ / $3\times10^{-5}$ \\
I5 & ViT-B/16-384 + AraBERT              & frozen    & $\dagger$ & $8 \times 2$ & fp16 & $5\times10^{-5}$ (matched, all arms) \\
\bottomrule
\end{tabular*}
\end{table*}

\label{sec:norm}

Every reference, prediction, and edition string passes through one normalizer: Unicode NFKC,
tatweel removal, diacritic stripping, whitespace collapse, and the orthographic unification of
alef and hamza forms in Table~\ref{tab:norm}. That unification is a real choice, not a formality.
It removes a class of orthographic variation that is genuinely ambiguous in historical hands, and
it lowers CER by several points relative to diacritic stripping alone. We report raw and
normalized scores side by side so the size of that effect stays visible. Because normalization
moves Arabic CER by whole points, we fix the policy and apply it uniformly; every table shows both
raw and normalized numbers, and no figure mixes policies. Exploratory reading of the Muharaf
transcriptions turned up accented Latin characters and other non-Arabic glyphs in the reference
text, which we keep rather than silently delete. That is one more reason raw and normalized scores
are reported apart.

\begin{table}[t]
\centering
\caption{Orthographic unification applied by the normalizer (Sec.~\ref{sec:norm}); NFKC, tatweel
removal, diacritic stripping, and whitespace collapse are applied in addition, identically to
references and predictions. Arabic glyphs are shown in transliteration for pdf\LaTeX{}
compatibility; a Xe\LaTeX{} variant with native script is available.}
\label{tab:norm}
\footnotesize
\begin{tabular*}{\linewidth}{@{\extracolsep{\fill}}ll@{}}
\toprule
Source form(s) & Normalized to \\
\midrule
Alef with hamza, madda, or wasl & bare alef \\
Alef maqsura   & ya \\
Ta marbuta     & ha \\
Waw-hamza      & waw (bare) \\
Ya-hamza       & ya (bare) \\
Standalone hamza & (removed) \\
\bottomrule
\end{tabular*}
\end{table}

\subsection{Scoring with uncertainty, and what an interval settles}
\label{sec:scoring}

CER and WER come from per-line Levenshtein edits, aggregated at corpus level:
\begin{equation}
\mathrm{CER} = \frac{\sum_i d_c(r_i, h_i)}{\sum_i |r_i|}, \quad
\mathrm{WER} = \frac{\sum_i d_w(r_i, h_i)}{\sum_i |r_i|_w}.
\end{equation}

Scores carry bootstrap 95\% confidence intervals [28] (1,000 resamples over lines; 2,000 for the
paired comparison in Section~\ref{sec:matched}). We claim that one reader beats another only
through paired-delta bootstrap intervals on shared lines, which respects the strong per-line
correlation between systems.

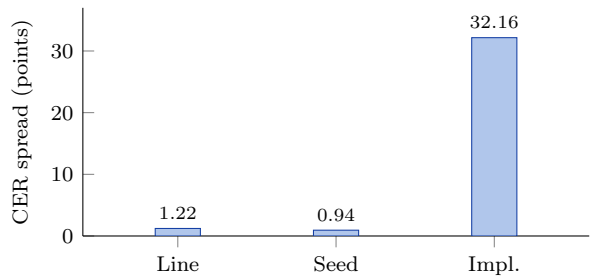
\begin{figure}[t]
\centering
\begin{tikzpicture}
\begin{axis}[
  width=\linewidth, height=4.6cm,
  ybar, bar width=17pt,
  ylabel={CER spread (points)},
  symbolic x coords={Line,Seed,Impl.},
  xtick=data,
  ymin=0, ymax=37,
  ytick={0,10,20,30},
  nodes near coords={\pgfmathprintnumber[fixed,fixed zerofill,precision=2]{\pgfplotspointmeta}},
  nodes near coords style={font=\scriptsize,text=black},
  every axis plot/.append style={fill=dalpale,draw=dalblue},
  tick label style={font=\footnotesize},
  label style={font=\footnotesize},
  axis lines*=left,
  enlarge x limits=0.30,
]
\addplot[fill=dalpale,draw=dalblue] coordinates {(Line,1.22) (Seed,0.94) (Impl.,32.16)};
\end{axis}
\end{tikzpicture}
\caption{What each level of uncertainty can see, on this study's data: the width of I1's paired
line-level interval (1.22), the seed-to-seed spread within I1's single-stage condition (0.94, the range across its
three seeds), and the range across the four implementations (32.16), all in CER points. The quantity that dominates
is the one standard reporting practice does not measure.}
\label{fig:levels}
\end{figure}

Three kinds of uncertainty need to be kept apart, and this paper exists because the third is
usually left out. \emph{Line level}: paired bootstrap over evaluation lines within one run, which
shows a gap is not an artifact of which lines were sampled. \emph{Seed level}: mean $\pm$ standard
deviation over seeds, which shows run-to-run reliability within one implementation.
\emph{Implementation level}: variation across separately configured runs of the same nominal
protocol, which shows whether the effect belongs to the method or to the implementer. A line-level
interval that excludes zero says nothing about the second or third. Where WER runs past 100\%,
that is expected rather than an error: word-level edits can exceed the reference word count when
the hypothesis is longer or heavily substituted, the hallucinatory failure mode of
Section~\ref{sec:zeroshot}. Figure~\ref{fig:levels} sets the three levels against each other on
this study's data; the line- and seed-level spreads are dwarfed by the implementation-level range.

\subsection{Readers}
\label{sec:readers}

Throughout, a \emph{reader} is the OCR model under evaluation, the system that transcribes a
manuscript line image, not the person reading this paper.

\textbf{Zero-shot floor.} Qwen2.5-VL-7B-Instruct [5] in 4-bit NF4 quantization, prompted in Arabic
for verbatim single-line transcription, with a per-sample generation cap scaled to reference
length. We state the quantization because a 4-bit model is a weaker floor than the full-precision
release; we report it as the floor a practitioner meets on a commodity GPU, not as the model's
ceiling.

\textbf{VLM-LoRA.} The same VLM adapted with LoRA [15] (rank 16, $\alpha = 32$, dropout 0.05, on
the q and v attention projections, vision tower frozen), trained staged KHATT$\rightarrow$Muharaf
with the loss masked to the assistant span.

\textbf{TrOCR family.} Four separately configured builds of the vision-encoder-decoder architecture
[10], each running the single-stage (M1) and staged (M2) conditions. Table~\ref{tab:impl} gives
their configurations. Two build on \texttt{RayR1/trocr-base-arabic-handwritten}, a community
checkpoint whose training data its model card does not disclose; we note in
Section~\ref{sec:limits} that overlap with Muharaf cannot be ruled out, so absolute CER for those
two may be optimistic. Two build a reader from a ViT encoder [11] and an AraBERT decoder [27] with
the encoder frozen throughout, a choice made after smoke tests showed that a randomly initialized
decoder destabilizes an unfrozen encoder.

\subsection{Preprocessing}

Line images share one pipeline: contrast-limited adaptive histogram equalization (CLAHE) for ink
and background separation across heterogeneous scans, Hough-transform deskewing, resizing to the
encoder's geometry with aspect-preserving padding, and a quality filter on height, aspect ratio,
and ink density that discards decorative elements, margins, and degenerate crops. Manifests are
integrity-checked so every image has a matching transcription, and image-text pairs are
spot-checked by eye before training.

\subsection{Compute-matched warm-up matrix (I5)}
\label{sec:matrix}

The four pilot runs disagree on the sign of the effect because they differ along several axes at
once. To answer the same question under control, we add one follow-up experiment in which the
warm-up corpus's domain is the only quantity that moves between arms.

\textbf{Reader.} To separate the data-schedule question from checkpoint provenance, every arm uses
one reader family trained from generic pretrained weights rather than an OCR-pretrained
checkpoint, so no arm inherits an undocumented prior exposure to KHATT or Muharaf. The encoder is
\texttt{google/vit-base-patch16-384} [11] (ImageNet-initialized); the decoder is
\texttt{aubmindlab/bert-base-arabertv02} [27], joined through Hugging Face's
\texttt{VisionEncoderDecoderModel}. Generation uses beam search (beam 4, no-repeat-3-gram, max
length 128). This is deliberately a smaller architecture than the pilots' OCR-pretrained
TrOCR-large: the matrix answers the pilots' question, it does not reproduce their absolute CER
(Sec.~\ref{sec:limits}).

\textbf{Design principle.} One rule governs the matrix: the only thing that changes between arms is
the warm-up data source. Four mechanisms enforce it. (i) Size-matched warm-up corpora: every
warm-up corpus is subsampled with a frozen seed to 9,497 lines, the size of the smallest available
corpus (KHATT), so no arm is matched upward. (ii) A fixed warm-up budget $N_w = 1{,}200$ steps for
every warm-up arm. (iii) A pilot-frozen target budget $N_t$, fixed once
(Sec.~\ref{sec:matrixsetup}) and held across all arms and seeds, so no arm buys extra target
training to make up for a worse start. (iv) Identical optimizer settings for every arm, stage, and
seed; no arm gets the stage-specific learning-rate cut that confounded the pilot
(Sec.~\ref{sec:fourimpl}). Reporting these settings for every arm, not only the one under
scrutiny, is a hard requirement for any $\Delta$CER claim, and the pilot's single most actionable
lesson.

\textbf{Arms.} The matrix has a same-domain compute-matched control (A+, an equal-size disjoint
slice of the Muharaf target data), the primary contrast (B, KHATT), a no-warm-up practitioner
reference (A0), and a best-case in-domain-real control (E, a disjoint Muharaf slice carved out
before the target split is formed). A0 is not compute-matched and is flagged as such on every run:
it skips the warm-up stage and so receives strictly fewer optimization steps than A+, B, or E. We
keep it as a deployment-relevant reference, but it is never the control and never enters the
primary or corrected contrasts as if it were.

\subsection{Domain-distance measurement}
\label{sec:distance}

To give domain gap a number rather than an intuition, each warm-up corpus is compared to the target
split on two independent axes, both computed from the frozen, pre-fine-tuning base encoder so the
distance axis cannot be contaminated by any arm's own training. On the image side we compute an
unbiased RBF-kernel Maximum Mean Discrepancy (MMD$^2$) [29] between base-encoder embeddings of up
to 2,000 images per corpus, with kernel bandwidth from the median pairwise-distance heuristic and a
200-fold bootstrap for a 95\% interval. On the text side we compute the Jensen-Shannon divergence
between character 3-gram frequency distributions, evaluated after the normalizer of
Sec.~\ref{sec:norm} so the text metric and the CER metric see identical text.

\subsection{Statistical protocol}

One contrast is pre-registered as primary and reported uncorrected: B vs.\ A+, KHATT warm-up
against the compute-matched, no-domain-shift control. The remaining contrasts (A0 vs.\ A+, E vs.\
A+) are secondary, each tested with a paired $t$-test and jointly corrected by Holm-Bonferroni at
$\alpha = 0.05$ [30], so adding secondary arms does not inflate the false-positive rate. Seed-level
contrasts use a paired Student-$t$ interval over per-seed CER differences ($n = 3$ seeds). We do
not test whether $\Delta$CER rises monotonically with measured domain distance: a Spearman ordering
test needs $\geq 3$ distance levels, and this study measures two (B and E).

\subsection{Edition alignment: a protocol for corpora without line-level labels}
\label{sec:edition}

Many heritage manuscript corpora have published scholarly editions, wording that experts have
verified, missing only the correspondence to specific page lines. We specify and release a protocol
that exploits this. A seed reader transcribes each line crop; the prediction is matched into the
normalized edition by boundary-aware partial-ratio alignment (score threshold 82, minimum
prediction length 6); and the adopted label is the edition's wording over the aligned span, never
the reader's own output. This boundary-correct formulation repairs the failure mode of naive prefix
matching and anchors label quality to expert text rather than model text. Reliability is measured
rather than assumed: a random sample of 100 aligned lines is judged correct or incorrect by a human
auditor, and precision is reported with a Wilson interval; anchors are trusted for adaptation only
if the audited precision supports it. For gold subsets, experts post-edit reader predictions rather
than transcribe from scratch, and correction time is logged.

\textbf{Status.} The protocol is specified, implemented, and released as part of
SaudiHeritage-OCR. It has not been run at scale, because acquisition of the target Saudi manuscript
corpus was not completed within this study. We therefore report no audited precision, and we
describe no labels here as verified in the measured sense the protocol defines. That would be
exactly the kind of unbacked claim this paper argues against.

\section{Experimental Setup}

Experiments ran on Google Colab GPU runtimes (NVIDIA L4, A100, T4) and on local RTX 4070-series
GPUs, with the detected device recorded per run. Dependencies are pinned (PyTorch, Transformers
$\geq 4.49$, PEFT, bitsandbytes, jiwer, rapidfuzz).

\textbf{VLM-LoRA lane.} Learning rate $10^{-4}$, 4-bit NF4, 3 epochs per stage, micro-batch 1 with
gradient accumulation 8 (effective batch 8), seeds $\{42, 43, 44\}$. Training used a 1,000-line
subset per stage and evaluation a 200-line sample, with 1,000-resample bootstrap intervals; this
lane is a controlled small-scale reader, not a scaled result, and is reported as such.

\textbf{TrOCR lanes.} Base learning rate $5 \times 10^{-5}$ except where Table~\ref{tab:impl}
states otherwise, beam width 4, max target length 128 tokens. I1 evaluated M1 with seeds
$\{42, 43, 44\}$ and M2 with $\{42, 43\}$; the remaining implementations report single runs.

\textbf{A deviation we report rather than hide.} Earlier drafts selected, per condition, the run
with the lowest normalized CER. That is a best-of-$n$ selection and it biases point estimates.
Table~\ref{tab:central} reports means over completed seeds; where only one run exists, the table
says so.

\subsection{Compute-matched matrix (I5) setup}
\label{sec:matrixsetup}

The matrix was trained on a single workstation GPU (NVIDIA RTX 4070 SUPER, 12\,GB) under PyTorch
2.5.1+cu121 and Hugging Face Transformers, fp16 throughout. Table~\ref{tab:splits} gives the data
splits, with the arm-E warm-up slice carved once from Muharaf train and disjoint from the target
split every arm shares; Table~\ref{tab:arms} defines the four arms and their warm-up sources;
Table~\ref{tab:hyper} lists the training hyperparameters. These settings are identical across every
arm and seed. That uniformity is the design's central control, not an incidental detail, and is
reported in full for that reason. The configuration is row I5 in Table~\ref{tab:impl}, beside the
four pilots.

\begin{table}[t]
\centering
\caption{Matrix data splits. The arm-E warm-up slice is carved from Muharaf train once, before
training, disjoint from the target split.}
\label{tab:splits}
\footnotesize
\begin{tabular*}{\linewidth}{@{\extracolsep{\fill}}lr@{}}
\toprule
Split & Lines \\
\midrule
Muharaf train (raw) & 22,091 \\
Muharaf validation  & 1,069 \\
Muharaf test        & 1,334 \\
\quad$\hookrightarrow$ Arm-E warm-up slice (from train, seed $= 7$) & 9,497 \\
\quad$\hookrightarrow$ Target split (remaining train, all arms) & 12,594 \\
Overlap, target split vs.\ E slice & 0 \\
\bottomrule
\end{tabular*}
\end{table}

\begin{table}[t]
\centering
\caption{Matrix arms. A+ is the compute-matched control: it spends the warm-up budget on additional
target-domain Muharaf passes before the common target stage. E uses a separately carved, disjoint
real-Muharaf slice, an in-domain data-source control. A0 omits the warm-up budget, is therefore not
compute-matched, and is never the primary control (Sec.~\ref{sec:matrix}).}
\label{tab:arms}
\footnotesize
\begin{tabular*}{\linewidth}{@{\extracolsep{\fill}}lll@{}}
\toprule
Arm & Warm-up source & Role \\
\midrule
A+ & Same-domain Muharaf   & Compute-matched ctrl. \\
B  & KHATT (modern hw)     & Primary contrast \\
A0 & None (target-only)    & Practitioner ref. \\
E  & Disjoint real-Muharaf & In-domain-real ctrl. \\
\bottomrule
\end{tabular*}
\end{table}

\begin{table}[t]
\centering
\caption{Matrix training hyperparameters, identical across every arm and seed.}
\label{tab:hyper}
\footnotesize
\begin{tabular*}{\linewidth}{@{\extracolsep{\fill}}ll@{}}
\toprule
Setting & Value \\
\midrule
Learning rate & $5 \times 10^{-5}$ \\
LR warmup ratio & 0.05 \\
Weight decay & 0.01 \\
Per-device batch size & 8 \\
Gradient accumulation & 2 (eff.\ batch 16) \\
Warm-up steps $N_w$ (A+, B, E) & 1,200 \\
Target steps $N_t$ (all) & 3,000 (pilot-frozen) \\
Warm-up corpus (A+, B, E) & 9,497 lines (matched) \\
Seeds & 42, 43, 44 ($n = 3$/arm) \\
Decoding & Beam 4, len 128, no-rep-3g \\
Precision & fp16 \\
Eval / checkpoint interval & every 500 steps \\
\bottomrule
\end{tabular*}
\end{table}

\textbf{Freezing the target-step budget $N_t$.} $N_t$ is not chosen per arm; it is fixed once,
before any arm or seed runs, from a single pilot (seed 42, A+-style target-only training, capped at
2,500 steps for the time budget), then frozen for every arm and seed. Validation loss fell
$5.521 \rightarrow 5.239 \rightarrow 5.095 \rightarrow 5.043 \rightarrow 5.042$ at steps 500 to
2,500; with best step $s^* = 2{,}500$, $N_t = \lceil s^* \cdot 1.2/500 \rceil \cdot 500 = 3{,}000$.
We state the limitation plainly: loss was still edging down at the 2,500-step cutoff, so
$N_t = 3{,}000$ is time-boxed rather than fully plateaued. Because $N_t$ is frozen identically
across the matrix, this biases no contrast, but each arm's absolute CER may sit at a
still-improving point on its curve (Sec.~\ref{sec:limits}).

\textbf{Evaluation.} Every arm and seed is evaluated once, at the end of its fixed target-step
budget, on the full 1,334-line Muharaf test set, with no early stopping on the test set and no
best-of-$n$ over checkpoints or seeds. CER and WER use the single shared normalizer
(Sec.~\ref{sec:norm}); WER is measured on normalized, whitespace-tokenized words.

\section{Results}

\subsection{Zero-shot floor, measured twice}
\label{sec:zeroshot}

The zero-shot floor is effectively unusable on this material (Table~\ref{tab:zeroshot}).
Predictions are fluent, heavily diacritized Arabic largely unrelated to the source: hallucination
rather than reading, which is why WER exceeds 100\%.

\begin{table}[t]
\centering
\caption{Zero-shot Qwen2.5-VL-7B-Instruct (4-bit NF4), 200-line Muharaf test samples. Two runs
differing only in generation control: A caps generation at $\min(2\times$ reference length, 80
tokens$)$ with repetition penalty 1.3 and no-repeat-3-gram; B caps at $2\times$ reference length
with repetition penalty 1.05.}
\label{tab:zeroshot}
\footnotesize
\begin{tabular*}{\linewidth}{@{\extracolsep{\fill}}lcccc@{}}
\toprule
 & \multicolumn{2}{c}{CER (\%)} & \multicolumn{2}{c}{WER (\%)} \\
\cmidrule(lr){2-3}\cmidrule(lr){4-5}
Run & raw & norm. & raw & norm. \\
\midrule
A & 84.46 & 78.59 & 113.85 & 111.39 \\
B & 95.14 & 80.71 & 114.68 & 111.42 \\
\bottomrule
\end{tabular*}
\end{table}

Two runs of the same nominal configuration, differing only in generation control, differ by 2.1
normalized CER points and 10.7 raw points. That is a clean secondary example of evaluation
sensitivity arising outside model training, and a calibration of how much of a reported OCR
difference can come from generation control alone.

On the single held-out Al-Mahd Hijazi line the floor produces 60.71\% CER / 83.33\% WER; the output
is well-formed Arabic whose content diverges from the reference. With $N = 1$ and a post-edited
gold reading, this is illustrative only.

\subsection{VLM-LoRA reader}
\label{sec:vlmlora}

\begin{figure}[t]
\centering
\begin{tikzpicture}
\begin{axis}[
  width=\linewidth, height=4.6cm,
  ylabel={Normalized CER (\%)},
  symbolic x coords={seed 42,seed 43,seed 44},
  xtick=data,
  ymin=41.5, ymax=50.5,
  ytick={42,44,46,48,50},
  tick label style={font=\footnotesize},
  label style={font=\footnotesize},
  axis lines*=left,
  enlarge x limits=0.30,
]
\draw[densely dashed,dalink] (axis cs:seed 42,45.00) -- (axis cs:seed 44,45.00);
\addplot[only marks,mark=*,mark size=2.2pt,color=dalblue,
  error bars/.cd, y dir=both, y explicit]
  coordinates {
    (seed 42,45.62) +- (0,3.66)
    (seed 43,44.29) +- (0,2.59)
    (seed 44,45.09) +- (0,3.11)
  };
\end{axis}
\end{tikzpicture}
\caption{VLM-LoRA on Muharaf across three seeds. The seed-to-seed spread ($\pm 0.55$) is far
smaller than the line-level bootstrap width ($\approx 5$ points). Within one implementation the
experiment behaves as the protocol assumes, which is why the instability in
Fig.~\ref{fig:central} cannot be dismissed as ordinary noise. Dashed line: seed mean
$45.00 \pm 0.55$; bars: line-level bootstrap 95\% CI.}
\label{fig:lora}
\end{figure}
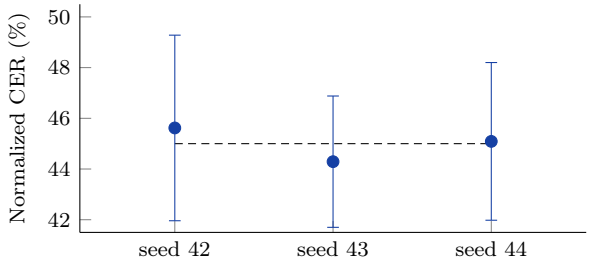

Table~\ref{tab:lora} reports the only three-seed reader in this study with per-seed bootstrap
intervals. Per-seed normalized CER on Muharaf was 45.62 / 44.29 / 45.09, with bootstrap 95\%
intervals $[42.65, 49.28]$, $[41.96, 46.88]$, $[42.38, 48.20]$. The seed-to-seed spread
($\pm 0.55$) is an order of magnitude smaller than the line-level interval width ($\approx 5$
points), which is the expected relationship and a useful contrast with Section~\ref{sec:matched}:
within one implementation this experiment is stable (Fig.~\ref{fig:lora}). The two stages are
evaluated on different corpora and are not comparable to one another; they document the reader, not
a schedule claim.

\begin{table}[t]
\centering
\caption{Qwen2.5-VL + LoRA, staged KHATT$\rightarrow$Muharaf, three seeds. Normalized scores, mean
$\pm$ SD over seeds (population SD, $n=3$). 1,000-line training subset per stage, 200-line evaluation sample.}
\label{tab:lora}
\footnotesize
\begin{tabular*}{\linewidth}{@{\extracolsep{\fill}}lccc@{}}
\toprule
Stage & Eval set & CER (\%) & WER (\%) \\
\midrule
1 (KHATT)   & KHATT val    & $32.53 \pm 0.77$ & $75.36 \pm 0.76$ \\
2 (Muharaf) & Muharaf test & $45.00 \pm 0.55$ & $81.76 \pm 1.12$ \\
\bottomrule
\end{tabular*}
\end{table}

\subsection{The data-schedule ablation across four implementations}
\label{sec:fourimpl}

Table~\ref{tab:central} and Fig.~\ref{fig:central} are the central result. Four runs of the same
nominal ablation produce effects spanning 32.16 CER points, with two reporting strong effects in
opposite directions.

\begin{table}[t]
\centering
\caption{Normalized CER on the Muharaf test split for the single-stage (M1) and staged
KHATT$\rightarrow$Muharaf (M2) conditions, across four implementations. $\Delta$ is M2 $-$ M1;
negative means the warm-up helped. Rows are sorted by $\Delta$CER, matching
Fig.~\ref{fig:central}. I1 reports means over 3 and 2 seeds respectively; the others are single
runs. The sign is not stable, and the two extremes are the two confounded runs.}
\label{tab:central}
\footnotesize
\begin{tabular*}{\linewidth}{@{\extracolsep{\fill}}lrrrl@{}}
\toprule
 & M1 & M2 & $\Delta$CER & Warm-up \\
\midrule
I2 & 45.86 & 28.22 & $-17.64$ & helps (contam.?) \\
I3 & 82.88 & 82.63 & $-0.25$  & no effect \\
I4 & 77.51 & 78.45 & $+0.94$  & no effect \\
I1 & 42.81 & 57.33 & $+14.52$ & hurts (LR confound) \\
\midrule
\multicolumn{5}{l}{Range across impl.\ 32.16 points, sign flips} \\
\bottomrule
\end{tabular*}
\end{table}

Inside I1 the paired line-level interval is narrow and excludes zero. A paired bootstrap on
identical held-out lines (2,000 resamples) gives $\Delta$CER $= +14.58$ points, 95\% CI
$[+13.97, +15.19]$, and the direction reproduces across both of I1's completed seeds (gaps of 14.65
and 14.88 points). The mean-over-seeds estimate in Table~\ref{tab:central} is $+14.52$; the two
figures are computed differently (paired-line bootstrap vs.\ mean of per-seed differences) and
reported as such. On its own, I1 would look like strong evidence of negative transfer. It is
contradicted, by a wider margin, by I2, which uses the same base checkpoint and the same corpora.

\textbf{Candidate drivers.} The runs differ along several axes at once (Table~\ref{tab:impl}), so
we cannot pin the divergence on one cause; we list the leading candidates in order of how
concretely each is evidenced.

\emph{Target-stage learning rate.} I1 is the only run whose two conditions do not share a
target-stage learning rate: M1 trains the Muharaf stage at $5 \times 10^{-5}$ while M2 trains it at
$1 \times 10^{-5}$, a deliberate reduction adopted to keep the staged run from forgetting. At a
matched 5-epoch budget on a reader far from convergence, a fivefold cut in learning rate reduces
adaptation to the target regardless of what came before. I1's penalty is therefore not cleanly
attributable to the warm-up, and I1 is the run reporting the largest penalty. This is the most
actionable finding in the paper: a stabilization choice implementers make routinely, and rarely
report, can exceed the effect under study.

\emph{Operating point.} I2 and I1 sit near 45\% and 43\% CER; I3 and I4 near 78 to 83\%. The two
runs reporting no effect are the two furthest from usable accuracy, where a schedule difference may
simply be undetectable.

\emph{Encoder freezing.} I3 and I4 freeze the encoder throughout; I1 and I2 do not. A frozen encoder
cannot adapt its visual representation to the warm-up domain, which plausibly suppresses both the
benefit and the harm of an out-of-domain stage, consistent with those two runs showing the smallest
effects. It also means their no effect should be read with care: part of the flatness may be the
design's inability to register the effect rather than the effect's absence.

\emph{Epoch budget.} I2 (8 epochs) and I1 (5 epochs) share a base checkpoint and differ in sign.
Whether I1's penalty would survive I2's longer budget is untested, and is a direct prediction of the
compute-matched experiment below.

\subsection{Compute matching shrinks the effect, and locates most of it outside the domain}
\label{sec:matched}

Under one controlled reader with every budget matched (Sec.~\ref{sec:matrix},
\ref{sec:matrixsetup}) the four arms fall in a narrow band (Table~\ref{tab:matrix}) and the primary
contrast is small and directionally consistent across the three shared seeds
(Table~\ref{tab:contrasts}). KHATT warm-up (B) is $+2.42$ CER points worse than the compute-matched
same-domain control (A+), with a paired seed-level 95\% interval of $[+0.60, +4.25]$. The effect is
consistent in direction across the three seeds and an order of magnitude smaller than the largest
pilot penalty. With $n = 3$ we read this as evidence for a small negative effect under this
configuration, not a universal claim.

\begin{table}[t]
\centering
\caption{Matrix per-arm normalized CER on the Muharaf test split, mean over $n = 3$ seeds with 95\%
CI. Absolute CER is high because the reader is trained from generic weights
(Sec.~\ref{sec:matrix}); the matrix measures the contrast, not the state of the art.}
\label{tab:matrix}
\footnotesize
\begin{tabular*}{\linewidth}{@{\extracolsep{\fill}}lrrc@{}}
\toprule
Arm & CER (\%) & SD & 95\% CI \\
\midrule
A+ (control)        & 79.16 & 1.44 & $[75.59, 82.74]$ \\
B (KHATT)           & 81.59 & 0.91 & $[79.32, 83.86]$ \\
A0 (no warm-up)     & 79.76 & 0.97 & $[77.36, 82.17]$ \\
E (in-domain real)  & 80.95 & 0.55 & $[79.58, 82.33]$ \\
\bottomrule
\end{tabular*}
\end{table}

\begin{table}[t]
\centering
\caption{Matrix contrasts vs.\ the control A+, seed-level paired interval. Positive $\Delta$ means
the warm-up hurt. Seed-level means differ from the difference of the per-arm means in
Table~\ref{tab:matrix} in the third decimal; $+2.42$ is the mean of the per-seed differences.}
\label{tab:contrasts}
\footnotesize
\begin{tabular*}{\linewidth}{@{\extracolsep{\fill}}lrcl@{}}
\toprule
Contrast & $\Delta$CER & 95\% CI & Role \\
\midrule
B $-$ A+  & $+2.42$ & $[+0.60, +4.25]$ & Primary contrast \\
A0 $-$ A+ & $+0.60$ & $[-0.84, +2.04]$ & Secondary (incl.\ 0) \\
E $-$ A+  & $+1.79$ & $[-3.10, +6.68]$ & Secondary (incl.\ 0) \\
B $-$ E   & $+0.64$ & (not tested)     & Domain-isolating \\
\bottomrule
\end{tabular*}
\end{table}

\begin{table}[t]
\centering
\caption{Measured domain distance to the target split (Sec.~\ref{sec:distance}). B's gap is large
on both axes; E's is near zero, a sanity check on the metric. The unbiased MMD$^2$ estimator is not
constrained to be positive, so E's point estimate can fall below the bootstrap interval of the
resampled statistic; both are reported as computed.}
\label{tab:distance}
\footnotesize
\begin{tabular*}{\linewidth}{@{\extracolsep{\fill}}lcc@{}}
\toprule
Arm & MMD$^2$ image [95\% CI] & JS (3-gram) \\
\midrule
B (KHATT)          & 0.4361 $[0.4263, 0.4429]$ & 0.2530 \\
E (Muharaf slice)  & $\approx$0.0000 $[0.0003, 0.0010]$ & 0.0167 \\
\bottomrule
\end{tabular*}
\end{table}

Two facts temper it, and we state them because the reader would find them anyway. First, the matrix
is noisy at this scale: the in-domain-real arm (E) sits at $+1.79$ over A+ with an interval of
$[-3.10, +6.68]$, about ten points wide on three seeds, and the no-warm-up reference (A0) is
$+0.60$ over A+ with an interval that also includes zero. Neither secondary comparison supports the
primary claim, and E's width is a caution on how precisely anything here can be resolved. Second,
and more important for interpretation: A+ and E are both Muharaf-derived warm-ups, yet they differ
by 1.79 points. A+ spends its warm-up budget on additional passes that overlap the target data,
while E uses a disjoint slice; the gap between them is therefore a data-source effect within the
target domain, not a domain effect. That matters for how B should be read. Against A+, KHATT is
$+2.42$ worse; against E, a disjoint slice of real Muharaf, the point-estimate gap is only
$81.59 - 80.95 = +0.64$ CER points. We did not compute a paired interval for B vs.\ E, so we do not
test it, but the arithmetic is clear: most of the $+2.42$ is shared with a real-target warm-up and
is not specific to KHATT's foreign domain. B vs.\ A+ remains the pre-registered primary contrast and
we report it as such; we simply decline to dress a 0.6-point domain gap as a general finding about
modern handwriting.

B is, by a wide margin, the arm with the largest measured domain distance from the target
(Table~\ref{tab:distance}), the qualitative shape a distance-predicts-penalty account would expect,
though with only two measured distance levels that ordering is not formally tested.

As a supplementary check, the within-seed line-level bootstrap for B vs.\ A+ gives $+3.09$, $+2.54$,
and $+1.64$ points for seeds 42 to 44, systematically narrower than the seed-level interval above.
That is exactly the gap that made the pilot's implementation-level instability invisible when only a
line-level interval was reported.

\subsection{Qualitative error analysis}
\label{sec:qualitative}

\begin{figure*}[!t]
\centering
\includegraphics[width=0.86\textwidth]{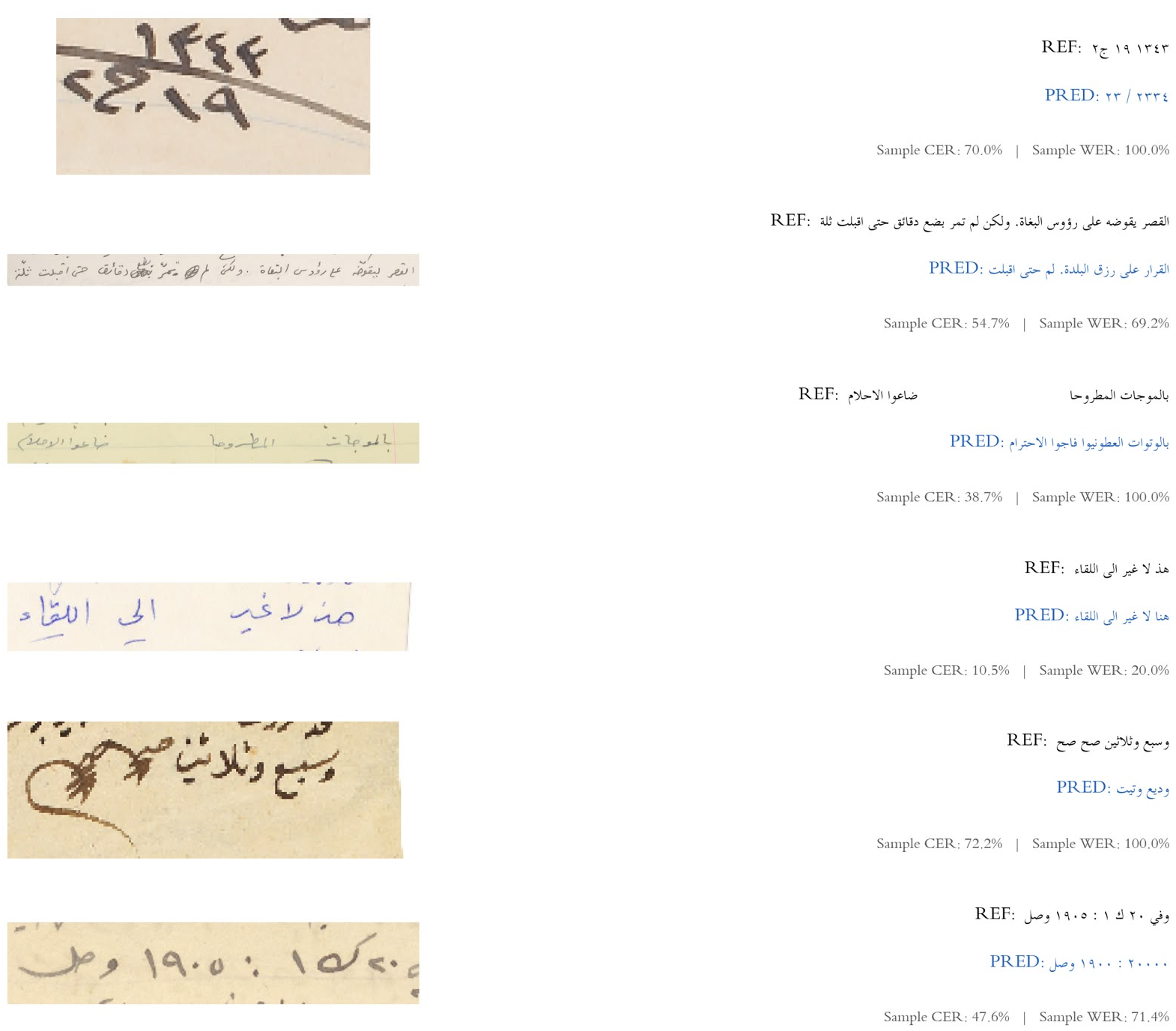}
\caption{Qualitative sample predictions from I1's single-stage (M1) Muharaf checkpoint. Each row
shows the manuscript line crop, the normalized reference (REF), the model prediction (PRED), and
per-sample CER / WER. The six lines are a fixed test sample (seed 7) shared across every checkpoint
in the study. Note the range within one model: 10.5\% CER on a clean modern-looking hand against
72.2\% on an ornate historical one, and WER at 100\% on lines whose CER is under 40\%, where the
reader recovers characters but not whole words.}
\label{fig:gallery}
\end{figure*}

Corpus-level CER says how much a reader is wrong; it does not say how. Figure~\ref{fig:gallery}
shows six Muharaf test lines transcribed by I1's single-stage checkpoint. The six lines are a fixed
sample (seed 7) reused across every trained checkpoint in the project so galleries are directly
comparable between lanes and seeds; they are an illustration of failure modes, not a sample estimate
of corpus CER, and the absolute levels carry I1's provenance caveat (Sec.~\ref{sec:limits}).

Three patterns are visible. First, the spread within one checkpoint is wide and tracks the hand
rather than the line length: the cleanest, most modern-looking hand in the sample is read almost
correctly at 10.5\% CER with a single substituted word, while an ornate decorated hand on the same
page stock collapses to 72.2\%. A single corpus CER averages over material that the reader handles
very differently. Second, numeral-dense lines fail in a specific way. On both dated lines the
surrounding words survive while the digit strings are replaced by plausible but wrong digits,
which suggests that numerals are undertrained relative to letters in the target corpus and that
date-bearing lines, exactly the lines an archival index would want, are the least reliable output.
Third, and most relevant to how the metrics in this paper should be read, WER reaches 100\% on
lines whose CER is well under half. The reader recovers much of the character sequence but rarely
lands a whole word intact, which is the mechanism behind the large CER-to-WER gap throughout our
tables and a reason to prefer CER as the primary metric on this material.

Across all six, the failure mode is substitution rather than collapse: the model emits fluent
Arabic in the register of the source document rather than empty or degenerate output. That
matches the hallucination pattern of the zero-shot floor (Sec.~\ref{sec:zeroshot}) in kind, though
not in degree, and it is why error rates on this material must be read alongside qualitative
inspection rather than on their own.

\subsection{Context against prior work}

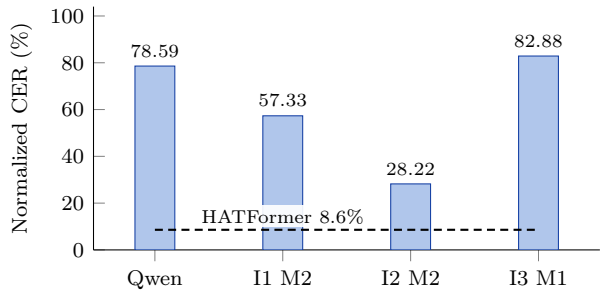
\begin{figure}[t]
\centering
\begin{tikzpicture}
\begin{axis}[
  width=\linewidth, height=4.8cm,
  ybar, bar width=15pt,
  ylabel={Normalized CER (\%)},
  symbolic x coords={Qwen,I1 M2,I2 M2,I3 M1},
  xtick=data,
  ymin=0, ymax=104,
  ytick={0,20,40,60,80,100},
  nodes near coords,
  nodes near coords style={font=\scriptsize,text=black},
  every axis plot/.append style={fill=dalpale,draw=dalblue},
  tick label style={font=\footnotesize},
  label style={font=\footnotesize},
  axis lines*=left,
  enlarge x limits=0.16,
]
\addplot[fill=dalpale,draw=dalblue] coordinates {(Qwen,78.59) (I1 M2,57.33) (I2 M2,28.22) (I3 M1,82.88)};
\draw[dalink,thick,densely dashed] (axis cs:Qwen,8.6) -- (axis cs:I3 M1,8.6);
\node[dalink,font=\scriptsize,anchor=south,fill=white,fill opacity=0.92,text opacity=1,inner sep=1.2pt]
  at (axis cs:I1 M2,9.6) {HATFormer 8.6\%};
\end{axis}
\end{tikzpicture}
\caption{Readers under one normalization policy, against the published state of the art. The best
and worst TrOCR configurations differ by tens of CER points on the same benchmark; both are
configurations a reasonable implementer produced.}
\label{fig:context}
\end{figure}

Every configuration here stays well above HATFormer's 8.6\% CER on Muharaf [4]. These readers
instrument the protocol and expose implementation variance; they are not attempts at the state of
the art, and the comparison is contextual rather than direct, since normalization policies, training
budgets, split construction, and architectures all differ across studies. The zero-shot behavior is
consistent with KITAB-Bench [6]. Figure~\ref{fig:context} places every reader on one normalized axis
against the published state of the art.

\section{Discussion}

\subsection{Interpretation}

The clearest result of this study is about evidence, not about Arabic OCR. A schedule ablation in
this setting, run once, with seeds and paired bootstrap intervals and every statistical practice the
field currently expects, can yield a confident finding whose sign reverses under a different
reasonable implementation of the same protocol. The interval was not wrong; it answered a narrower
question than the claim it was used to support. Line-level intervals quantify which lines were
sampled. Seed-level statistics quantify run-to-run noise. Neither sees the implementer.

For practitioners adapting readers to Saudi heritage material, the operational conclusion is that
published single-implementation schedule results, including the earlier version of our own, should
not be used to allocate annotation and compute budgets. The broader question stays open across
architectures and training regimes. Within the controlled I5 configuration our data support a small
negative effect of KHATT warm-up, but, as Section~\ref{sec:matched} shows, the part specific to the
modern-handwriting domain is small once the right control is used. The cross-implementation results
further suggest that whatever effect exists can depend on the operating point and on adaptation
choices.

For the wider field, one mechanism is worth generalizing beyond Arabic HTR. Stabilization
adjustments to a staged run, such as lowering the learning rate, shortening the target stage, or
freezing a component, are made for good engineering reasons, are rarely reported as experimental
conditions, and are applied asymmetrically, to the staged arm and not to the baseline. That
asymmetry is a confound in the direction of the reported effect, and it is invisible in a results
table.

\subsection{What the protocol contributes}

The interval-first apparatus keeps its value; it is the reason the divergence in
Table~\ref{tab:central} is interpretable rather than merely noisy. A single normalizer is what
makes the four runs comparable at all; without it, a 2 to 5 point spread would be attributable to
normalization alone, as the raw-versus-normalized columns of Table~\ref{tab:zeroshot} make
concrete. Our own addition is the third uncertainty level, together with a recommendation: any
schedule claim in Arabic HTR should come either with an independent re-implementation or with an
explicit statement that implementation-level variance was not assessed.

\subsection{Limitations}
\label{sec:limits}

The four implementations were not designed as a controlled replication; they differ along several
axes at once, so Table~\ref{tab:central} establishes that the effect is unstable but cannot
decompose why. That is a limitation on the diagnosis, not on the finding. Two implementations build
on a community checkpoint whose training data its model card does not disclose; if that checkpoint
saw Muharaf, the test split is contaminated and I1's and I2's absolute CER is optimistic by an
unknown amount. We report this rather than drop those runs, because the instability they demonstrate
does not depend on their absolute level; but no absolute number from I1 or I2 should be cited as a
Muharaf result. Contamination threatens the absolute level more than the within-run $\Delta$CER,
since both conditions share the checkpoint, which is one reason we lean on the compute-matched
matrix for the directional claim. The implementations report different numbers of seeds (three,
two, and single runs), so seed- and implementation-level variance are partly entangled, and I4's
independent manifest and resplit make its M1 and M2 internally comparable but its absolute level not
comparable to the others. The VLM-LoRA lane trains on 1,000-line subsets and evaluates on 200-line
samples; it documents a reader and does not scale to a benchmark claim. The Muharaf public split
lacks writer identifiers, so writer-independence cannot be confirmed. The Al-Mahd result is a single
line with a post-edited gold reading, so no manuscript-to-stone gap is measured. The
edition-alignment protocol is released as method, with no audited precision, because target-corpus
acquisition was not completed. All readers operate at the line level; page layout, reading-order
reconstruction, and marginalia are out of scope.

A final set of caveats concerns the compute-matched matrix. Its reader (ViT-Base + AraBERT from
generic weights) differs from the pilots' OCR-pretrained TrOCR-large, so the matrix answers the
pilots' question but does not reproduce their absolute CER. $N_t$ is time-boxed, not plateaued; this
affects all arms equally and biases no contrast, but each arm's absolute CER may sit below its
ceiling. Each arm uses only $n = 3$ shared seeds, so the seed-level interval is fragile and should
be read descriptively. With only two measured distance levels (B and E) we do not run a
distance-vs-$\Delta$CER ordering test. And A0 is not compute-matched and must not be read as a
controlled zero-warm-up arm.

\subsection{Implications for Saudi heritage}

A historical Arabic HTR system that reached manuscript-level accuracy could turn digitized heritage
archives into searchable textual corpora, opening the way to large-scale indexing, retrieval, and
scholarly analysis. Within that goal, this study's contribution is mainly methodological. Rather
than recommending a training schedule, we show that an intermediate-training strategy that looks
helpful on a single reported ablation is not robust across implementations and can produce
contradictory conclusions, which argues against adopting such a strategy on the strength of isolated
results. To support more reliable evaluation of alternatives, we provide a unified normalization
policy, a scoring framework, a verified KHATT decoder, and reproducible manifests. The
edition-alignment protocol addresses a real data limitation in Saudi manuscript digitization, namely
that scholarly editions exist without the line-level ground truth that HTR development and
evaluation need, and is released as a reproducible artifact that gives a foundation for building
auditable line-level datasets once access to further Saudi collections becomes available.

\section{Conclusion and Future Work}

This study examined how reliable intermediate-domain training is for historical Arabic HTR, through
repeated implementation and controlled experimentation. Four separately configured runs of one
nominal transfer-learning ablation, evaluated under a common normalizer and interval-based scorer,
produced effects from $-17.64$ to $+14.52$ CER points and changed direction. We report the full
configuration matrix and identify an asymmetric cut in the target-stage learning rate, in the run
with the largest observed penalty, as the most concrete candidate explanation for the divergence;
the two runs without such a confound show essentially no effect. We then ran a compute-matched
experiment to see whether the instability persisted once the major training variables were
controlled. Across three shared seeds, KHATT warm-up was worse than the compute-matched same-domain
control by a mean of 2.42 CER points; measured against a disjoint slice of real target data, the
domain-specific part of that gap is only about 0.6 points, so we present the result as a small
negative effect under this configuration rather than as a general property of modern-handwriting
warm-up. All conditions used identical warm-up and target-stage step budgets, an identical
target-stage learning rate, and a warm-up corpus subsampled to the same number of lines; the only
systematic difference in the warm-up stage was the domain of the corpus.

The takeaway is that conclusions about intermediate-domain training in historical Arabic HTR can be
sensitive to implementation choices, and that confidence intervals computed within a single
implementation do not capture that source of variation. The resulting framework emphasizes compute
matching, multi-seed evaluation, consistent preprocessing, and explicit reporting of
implementation-level variation when assessing transfer strategies for historical Arabic text
recognition.

Several directions remain. We will grow the Al-Mahd evaluation set to a statistically meaningful
sample and establish expert-from-scratch gold transcriptions, so Saudi epigraphic performance can be
measured without model-assisted post-editing. Once corpus access allows, we will run and audit the
edition-alignment protocol on larger Saudi manuscript collections to test its effectiveness for
building reliable line-level ground truth. And we will investigate restoration of damaged or
incomplete inscriptions along the lines of Ithaca [16], integrating recognition and restoration
while preserving uncertainty and expert oversight, moving the work beyond transcription toward a
broader framework for AI-assisted reading, reconstruction, and analysis of Saudi historical and
epigraphic heritage.

\section*{Code and Data Availability}

The SaudiHeritage-OCR package accompanying this paper is made publicly available upon
publication. It contains: the single normalization policy of Section~\ref{sec:norm} as an importable normalizer;
the CER/WER scorer with bootstrap and paired-delta confidence intervals and the Wilson-interval and
Holm-Bonferroni utilities used in Section~\ref{sec:scoring}; the verified KHATT v1.0 coded-label
decoder, which achieves 100\% code coverage on the training and validation CSVs of the commonly
mirrored release; the experimental manifests and split definitions for every implementation
reported here, including the compute-matched matrix of Section~\ref{sec:matrix}; the zero-shot and
LoRA VLM baseline scripts; and the edition-alignment protocol of Section~\ref{sec:edition} as a
runnable implementation together with its audit tooling.

Muharaf and KHATT are third-party corpora and are not redistributed; the package retrieves them
from their original sources and reproduces our splits from the released manifests. The Al-Mahd
inscription material is deliberately \emph{not} released as a benchmark. It is a single held-out
line with a post-edited reference reading, published epigraphic readings remain the property of the
originating survey and edition, and distributing one line as an evaluation set would invite exactly
the kind of underpowered claim this paper argues against. We will release a Saudi epigraphic
evaluation set when it reaches a statistically meaningful size with expert-from-scratch
transcriptions.

\section*{Author Contributions}

All authors contributed equally to this work. Roles below follow each author's own development
record.

\textbf{S.~Almarshad} established the shared setup and verified the evaluation pipeline, ran the
Qwen2.5-VL zero-shot baseline including the held-out Al-Mahd Hijazi inscription case, and carried
out the first TrOCR fine-tuning pipeline verification.
\textbf{M.~Alamri} supervised the project and advised on experimental design and manuscript
preparation.
\textbf{D.~Aloraini} built the KHATT coded-label decoder against the official v1.0 lookup table and
verified 100\% code coverage, ran implementations I2 and I3 and the three-seed VLM-LoRA lane, and
contributed the zero-shot run reported as A in Table~\ref{tab:zeroshot}.
\textbf{F.~Altuwaim} established the project structure and pinned dependencies, analysed the
candidate corpora, and implemented the download, verification, and preprocessing pipeline (CLAHE,
Hough deskewing, aspect-preserving padding, Arabic Unicode normalization) together with the data
loader.
\textbf{A.~AlOtaibi} built the shared scoring engine---the single normalizer, CER/WER, bootstrap and
paired-delta confidence intervals, and Wilson intervals---specified the edition-alignment protocol
of Section~\ref{sec:edition}, and led experiment orchestration, results aggregation, and manuscript
integration.
\textbf{R.~Alyabis} carried out exploratory analysis across the candidate corpora, built the Muharaf
manifest and split and the evaluation function with its test cases, contributed the zero-shot run
reported as B in Table~\ref{tab:zeroshot}, and ran implementation I4.
\textbf{R.~Aldawsari} ran implementation I1 across seeds in both the single-stage and staged
conditions, performed the paired significance testing on shared test lines, produced the
qualitative sample galleries of Section~\ref{sec:qualitative}, and built the step-based
checkpointing and crash-resilient training infrastructure used for the long local runs.

\section*{Acknowledgements}

We thank the creators and maintainers of the two public corpora this study depends on: the Muharaf
team for releasing expertly transcribed historical Arabic text lines, and the KHATT team for the
modern-handwriting database and the official v1.0 documentation whose lookup table made the
commonly mirrored coded-column release reproducibly usable. We thank the epigraphists whose
published readings of the Al-Mahd material made our out-of-domain qualitative case possible. This
work rests on open-source software, and we acknowledge the maintainers of PyTorch, Hugging Face
Transformers and PEFT, bitsandbytes, jiwer, and rapidfuzz.

\section*{Competing Interests}

The authors declare no competing interests.

\balance

\end{document}